\documentclass{article}

\usepackage{microtype}
\usepackage{graphicx}
\usepackage{booktabs} 

\usepackage{subfig}
\usepackage{times}
\usepackage{epsfig}
\usepackage{graphicx}
\usepackage{amsmath}
\usepackage{amssymb}
\usepackage{algorithm}
\usepackage{algorithmic}
\usepackage{tabularx}
\usepackage{enumitem}
\usepackage{amsmath}
\usepackage{booktabs}
\usepackage{makecell}
\usepackage{balance}

\usepackage{mathrsfs}
\usepackage{xcolor}

\usepackage{changes}
\usepackage{xspace}
\newcommand{\name}{\textit{few-shot NAS}\xspace}
\newcommand{\Name}{\textit{Few-shot NAS}\xspace}

\newcommand{\nasbench}{{NasBench-201}\xspace}
\newcommand{\nasbencho}{{NasBench-101}\xspace}

\newcommand{\cifara}{CIFAR-10\xspace}
\newcommand{\cifarb}{CIFAR-100\xspace}
\newcommand{\imagenet}{ImageNet-16-120\xspace}

\newcommand{\1}{{\em (i)}}
\newcommand{\2}{{\em (ii)}}

\usepackage{hyperref}

\usepackage[accepted]{icml2021}

\icmltitlerunning{Few-shot Neural Architecture Search}

\begin{document}

\twocolumn[
\icmltitle{Few-shot Neural Architecture Search}



\icmlsetsymbol{equal}{*}

\begin{icmlauthorlist}
\icmlauthor{Yiyang Zhao}{equal,wpi}
\icmlauthor{Linnan Wang}{equal,brown}
\icmlauthor{Yuandong Tian}{fb}
\icmlauthor{Rodrigo Fonseca}{brown}
\icmlauthor{Tian Guo}{wpi}
\end{icmlauthorlist}

\icmlaffiliation{wpi}{Worcester Polytechnic Institute}
\icmlaffiliation{brown}{Brown University}
\icmlaffiliation{fb}{Facebook AI Research}

\icmlcorrespondingauthor{Yiyang Zhao}{yzhao10@wpi.edu}

\icmlkeywords{Machine Learning, ICML}

\vskip 0.3in
]



\printAffiliationsAndNotice{\icmlEqualContribution} 

\begin{abstract}
Efficient evaluation of a network architecture drawn from a large search space remains a key challenge in Neural Architecture Search (NAS). Vanilla NAS evaluates each architecture by training from scratch, which gives the true performance but is extremely time-consuming.  Recently, one-shot NAS substantially reduces the computation cost by training only one supernetwork, a.k.a. \emph{supernet}, to approximate the performance of every architecture in the search space via weight-sharing. However, the performance estimation can be very inaccurate due to the co-adaption among operations~\cite{BG_understanding}. 
In this paper, we propose \emph{few-shot NAS} that uses multiple supernetworks, called \emph{sub-supernet}, each covering different regions of the search space to alleviate the undesired co-adaption. 
Compared to one-shot NAS, \emph{few-shot NAS} improves the accuracy of architecture evaluation with a small increase of evaluation cost. 
With only up to 7 sub-supernets, \emph{few-shot NAS} establishes new SoTAs: on ImageNet, it finds models that reach 80.5\% top-1 accuracy at 600 MB FLOPS and 77.5\% top-1 accuracy at 238 MFLOPS; on CIFAR10, it reaches 98.72\% top-1 accuracy without using extra data or transfer learning. 
In Auto-GAN, \emph{few-shot NAS} outperforms the previously published results by up to 20\%. Extensive experiments show that \emph{few-shot NAS} significantly improves various one-shot methods, including 4 gradient-based and 6 search-based methods on 3 different tasks in \nasbench and NasBench1-shot-1.

\end{abstract}

\section{Introduction}

\begin{figure}[t]
\centering 
  \begin{center}
    \includegraphics[width=0.9\columnwidth]{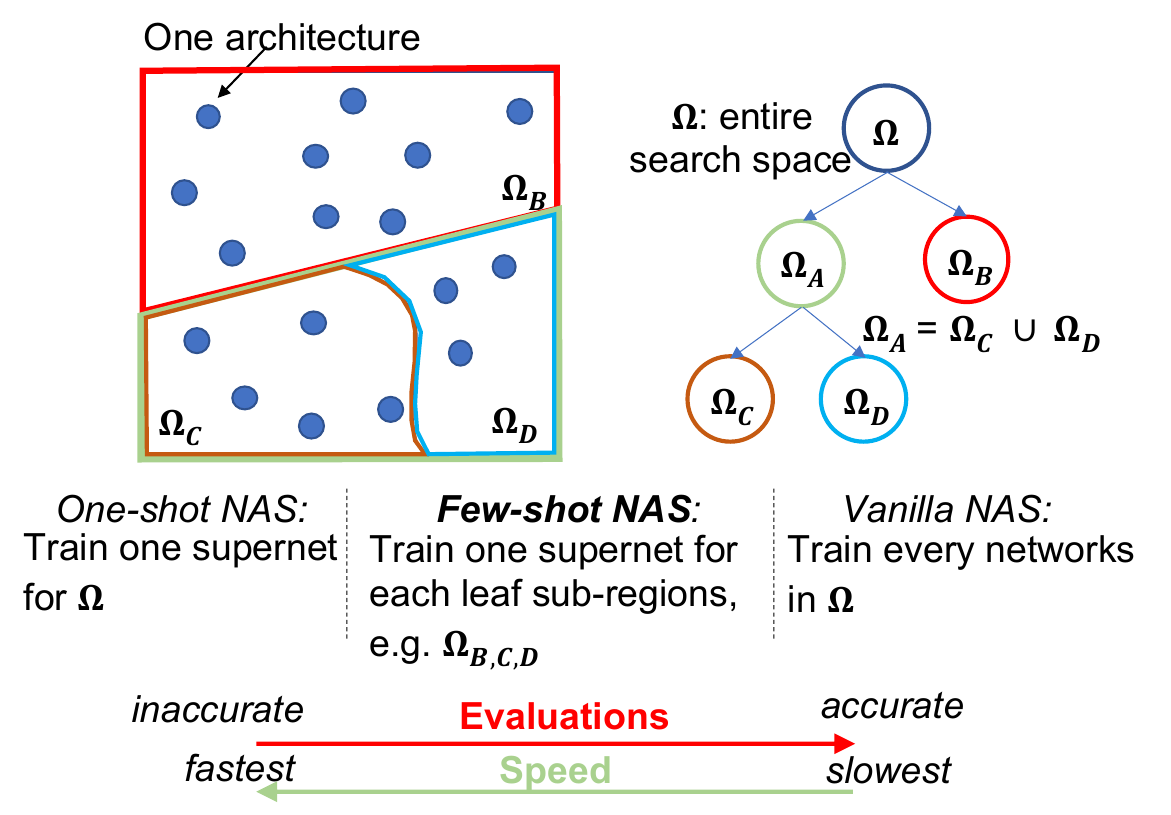}
  \end{center}
    \caption{Few-shot NAS is a tradeoff between the vanilla NAS and one-shot NAS that intends to maintain accurate evaluations in vanilla NAS and the speed advantages of one-shot NAS. }
    \label{fig:connection_one_standard_nas}
\end{figure}

\begin{figure}[t]
\centering 
  \begin{center}
    \includegraphics[width=\columnwidth]{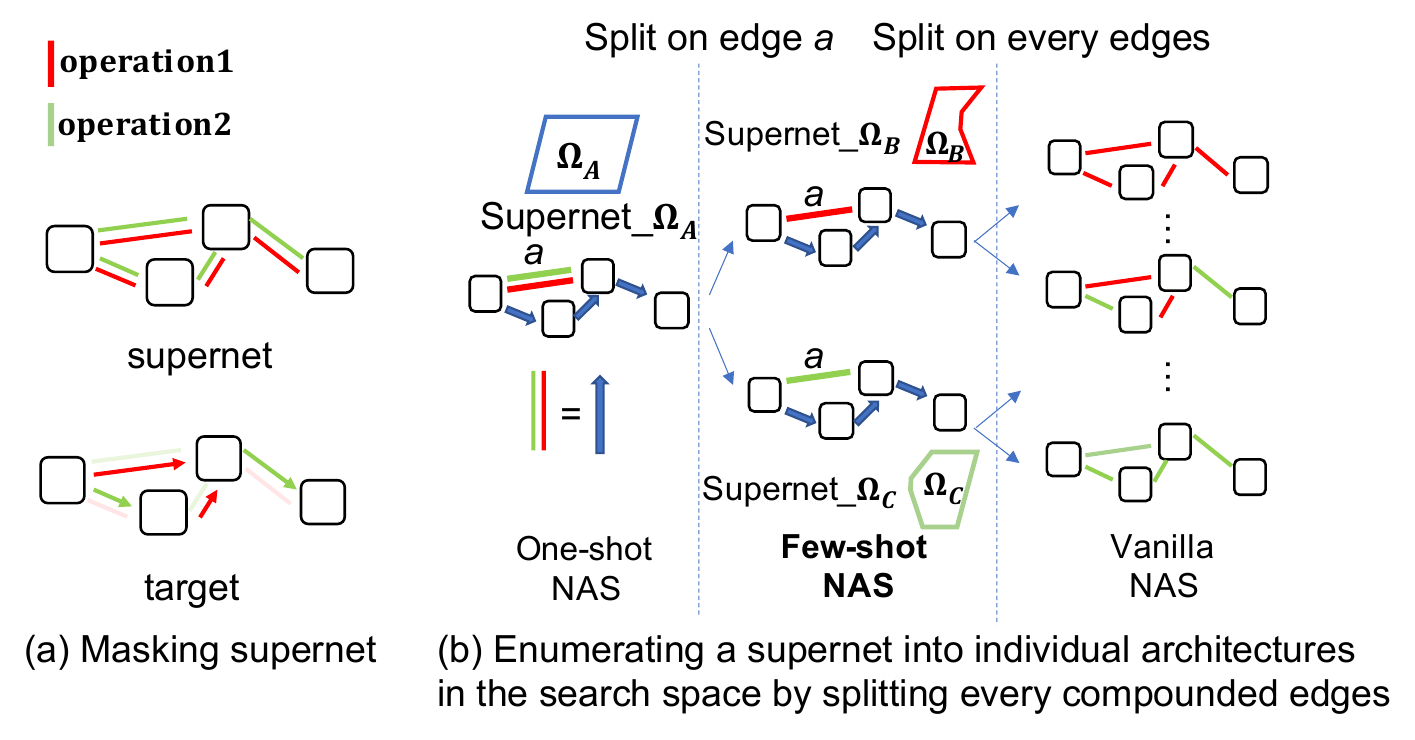}
  \end{center}
  \vspace{-0.5cm}
    \caption{(a) Masking supernet to a specific architecture for fast evaluation of network architecture. (b) the motivation of using few-shot NAS to alleviate the co-adaption impact. After splitting on edge $a$, supernet\_$\Omega_{B}$ exclusively predicts architectures in $\Omega_{B}$, so does supernet $\Omega_{C}$. }
    \label{fig:motivation}
\end{figure}

Neural Architecture Search (NAS) has attracted lots of interest over the past few years~\cite{nasnet,mnasnet,baker2017designing}. Using NAS, many deep learning tasks~\cite{detnas,autogan,autodeeplab,alphax,lanas} improve their performance without human tuning. Vanilla NAS requires a tremendous amount of computational costs (e.g., thousands of GPU hours) in order to find a superior neural architecture~\cite{nasnet,baker2017designing,rea}, most of which is due to evaluating new architecture proposals by training them from scratch. To reduce the cost, one-shot NAS~\cite{enas, DARTS} proposes to train a single supernet that represents all possible architectures in the search space. With supernet, the performance of a specific architecture can be approximately evaluated by picking the corresponding weights from the supernet (and masking out other missing edges, see Fig.~\ref{fig:motivation}(a)) without training, reducing the evaluation (and thus search) cost to just a few days (hours).

However, one-shot NAS suffers from degraded search performance due to inaccurate predictions from the supernet. Other works also have explicitly shown that using supernet degrades the final performance due to inaccurate performance predictions. For example, \cite{yu2019evaluating} observes that, without using the supernet, the average performance of NAS algorithms such as ENAS and NAO is 1\% higher than using it on NASBench-101, and they also conclude that the supernet never produces the true ranking. Besides, many works~\cite{BG_understanding,kaichengeva,nao,nasbench201,renqianBalance} also show that there is a non-trivial performance gap between the architectures found by one-shot NAS and vanilla NAS. Being consistent with the analysis in~\cite{yu2019evaluating}, the main reason for this is that the performance predicted by the supernet has a low correlation with the true performance. As an example, Section~\ref{sec:bg_motivation} shows that inaccurate performance prediction by supernet biases the search towards a wrong direction and hurts both the efficiency and the final results.

In this work, we propose \emph{few-shot NAS} that uses multiple supernets in the architecture search. While one supernet may not be able to model the entire search space due to its limited capacity and co-adaption of operations~\cite{BG_understanding}, using multiple supernets effectively addresses these issues by having each supernet modeling one part of the search space. The supernet allows parallel operations, i.e., \emph{compound edges}, shown in Fig.~\ref{fig:motivation}(b) as multiple lines of different colors. In this case, we separate the entire search space $\Omega$ (i.e., all possible network operations) into disjoint partitions by picking one edge from the compound edge and assign a \emph{sub-supernet} to model each leaf partition respectively. This procedure can be done in a recursive manner to yield a hierarchical partition. Although few-shot NAS increases the number of supernets, they can be trained efficiently by using a cascade of transfer learning: first, the root supernet is trained, then the child sub-supernet starts with the weights of the root and gets fine-tuned, etc. In this manner, we construct a collection of supernets, each of which is responsible for a region of the search space. 
Please refer to section~\ref{method} for the methodology details.

\begin{figure}[t]
\centering 
\includegraphics[width=1.0\columnwidth]{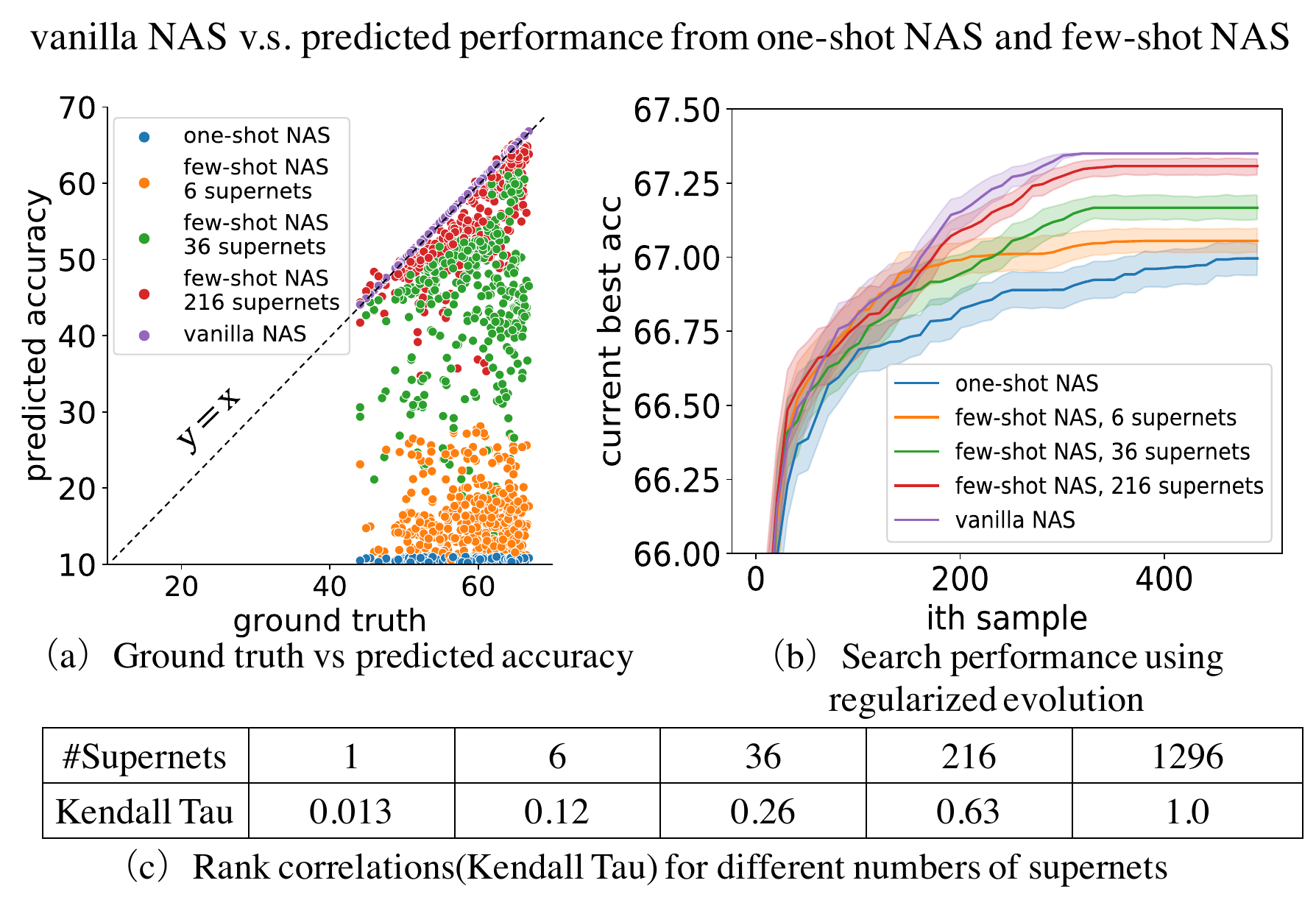}
\caption{(a) Using multi-supernets clearly improves the correlation and (c) provides the correlation score (Kendall Tau) at different numbers of supernets in (a); (b) shows the improved performance predictions result in better performance on NAS. }
\label{fig:validate-motivation}
\end{figure}

An immediate question is whether a lot of sub-supernets are needed to outperform one supernet. It turns out that a few sub-supernets already lead to strong performance, as demonstrated in Fig.~\ref{fig:validate-motivation}. Empirically, with only $5$ sub-supernets, we show that our \name greatly improved many existing NAS algorithms on \nasbench~\cite{nasbench201} and several popular deep learning tasks in Section~\ref{experiment}. Particularly, with our \name, we found SOTA efficient models that demonstrate 80.5\% top-1 accuracy at 600 MB FLOPS and 77.5\% top-1 accuracy at 238 MFLOPS on ImageNet, and 98.72\% top-1 accuracy on CIFAR-10 without using extra data or transferring weights from a network pre-trained on ImageNet. Moreover, by reusing the same search code from AUTOGAN~\cite{autogan}, \name also improved the results in~\cite{autogan}, from 12.42 to 10.73 in FID score.

\begin{table*}[t]
\caption{The definition of notations used through the paper.}
\resizebox{\textwidth}{!}{%
\begin{tabular}{@{}llllll@{}}
\toprule
$\Omega$                   & the whole architecture space                 & $\mathcal{A}$    & an architecture in the architecture space             & $m$                            & number of operations in the architecture space       \\
$\mathcal{S}$              & supernet                              & $N_{i}$          & the $i$th node in the architecture space              & $n$                            & number of nodes in the architecture space             \\
$\Omega^{'}$               & a sub-region of the whole architecture space & $E_{ij}$         & the compound edge between node $i$ and $j$ & $\mathcal{W}$                             & weights of neural network                    \\
$\mathcal{S}^{\Omega^{'}}$  & a sub-supernet                        & $f(\mathcal{A})$ & the evaluation of $\mathcal{A}$                 & $f(\mathcal{S}_{\mathcal{A}})$ & the evaluation of $\mathcal{A}$ by supernet \\ \bottomrule
\end{tabular}
}

\label{defination}
\end{table*}

\section{Background and Motivation}
\label{sec:bg_motivation}
The negative impact of co-adaption of operations from a compound edge was first identified by Bender et al.~\cite{BG_understanding}, and they show that the compound operations on the edge of the supernet can degrade the correlation between the estimated performance from a supernet and the true performance from training-from-scratch. While Bender et al. primarily focused on using drop path or dropout to ensure a robust supernet for performance prediction, our method was motivated by the following observation on one-shot NAS and vanilla NAS.

One-shot NAS uses a supernet to predict the performance of a specific architecture by deactivating the extra edges w.r.t a target architecture on the supernet via masking (Fig.~\ref{fig:motivation}(a)), then perform evaluations using the masked supernet. Therefore, we can view supernet as a representation of search space $\Omega$, and by masking, supernet can transform to any architectures in $\Omega$. This also implies we can enumerate all the architectures in $\Omega$ by recursively splitting every compound edge in a supernet. Fig.~\ref{fig:motivation}(b) illustrates the splitting process, the root is the supernet and leaves are individual architectures in the search space $\Omega$; the figure illustrates the case of splitting the compound edge $a$, and the recursively split follows similar procedures on all compound edges. In Fig.~\ref{fig:motivation}(b), one-shot NAS is the fastest but the most inaccurate in evaluations, while vanilla NAS is the most accurate in evaluations but the slowest. However, the middle ground, i.e., using multiple supernets, between one-shot NAS and vanilla NAS remains unexplored.

In a supernet, the effect of co-adaption results from combined operations on edges. Therefore, the evaluation of vanilla NAS is the most accurate. Based on this logic, it seems using several sub-supernets is a reasonable approach to alleviate the co-adaption effect by dissecting a compound edge into several separate sub-supernets that take charge of different sub-regions of the search space. For example, Fig.~\ref{fig:motivation}(b) shows few-shot NAS eliminates one compound edge $a$ after splitting, resulting in two supernets for $\Omega_{B}$ and $\Omega_{C}$, respectively. So, the predictions from the resulting sub-supernet are free from the co-adaption effects from the split compound edge $a$.

We designed a controlled experiment to verify the assumption that \emph{using multi-supernets improves the performance prediction}. First, we designed a search space that contains 1296 architectures and trained each architecture toward convergence to collect the final evaluation accuracy as the ground truth. Then we split the one-shot version of supernet into 6, 36, 216 sub-supernets following the procedures in Fig.~\ref{fig:motivation}(b). Finally, we trained each supernet with the same training pipeline in~\cite{BG_understanding}, and compared the predicted 1296 architecture performance to the ground truth using 1 (one-shot NAS), 6, 36, 216 supernets. Fig.~\ref{fig:validate-motivation} visualizes the results, and it indicates using multi-supernets significantly improves the correlation between predicted performance and the ground truth. Specifically, in Fig.~\ref{fig:validate-motivation}(c) the Kendall’s Tau~\cite{tau} ranking correlation of using 1 supernet (one-shot NAS), 6 supernets, 36 supernets, 216 supernets are 0.013, 0.12, 0.26, 0.63, respectively. As a result, the search algorithm takes fewer samples to find better networks due to more accurate performance predicted from supernets (Fig.~\ref{fig:validate-motivation}(b)).

In sec~\ref{experiment}, we conducted extensive experiments on various applications to ensure the proposed idea will generalize to other domains, including image recognition, language modeling, and image generation using Generative Adversarial Network (GAN). While all the experiments suggest few-shot NAS is an effective approach, we empirically have multiple supernets to train, and we propose a recursive fine-tuning work in sec.~\ref{method} that can reduce the supernet training time.

\section{Methodology}
\label{method}

In designing \name, we answer the following questions: \1 how to divide the search space represented by the one-shot model to sub-supernets and how to choose the number of sub-supernets given a search time budget (Section~\ref{subsec:split_strategies})?
\2 how to reduce the training time of multiple sub-supernets (Section~\ref{sub:transfer})? We also describe the integration of \name with existing NAS algorithms in Section~\ref{subsec:integration_gradient} and Section~\ref{subsec:integration_search_based}.

\subsection{Design of Split Strategy}
\label{subsec:split_strategies}

\begin{figure}[h]
\centering 
  \begin{center}
    \includegraphics[height=1.6cm]{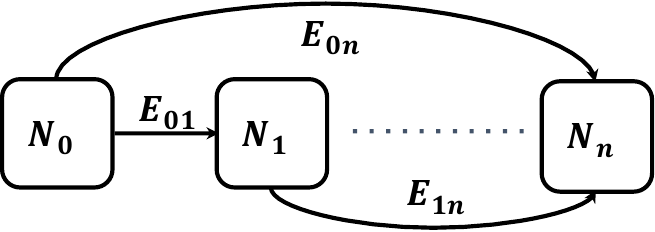}
  \end{center}
    \caption{A generic architecture space.}
    \label{fig:supernet} 
\end{figure}

Section~\ref{sec:bg_motivation} demonstrates the key insight of our method that the evaluation $f(\mathcal{S}^{\Omega^{k}}_{\mathcal{A}})$ of an architecture $A$ using a sub-supernet $\mathcal{S}^{\Omega^{k}}$ is more accurate as $\Omega^{k}$ gets smaller. However, the time to split the initial architecture space $\Omega$ grows exponentially with tree depth. Therefore, the ideal split is to find the best trade-off of split supernet numbers and the total search time.

\paragraph{Definition of a Generic NAS Space.}

Before we describe our split strategy, we first define a generic NAS space that is compatible with one-shot NAS. We use this architecture space to introduce some necessary concepts that will be used throughout the paper. 

The whole architecture space $\Omega$ is represented by a directed acyclic graph (DAG) shown in Figure~\ref{fig:supernet}. Each node denotes a latent state, e.g., feature maps in CNNs, and each compound edge represents a mixture of operations. We consider an architecture space with $n$ nodes and $m$ operations. Each node $i$ is denoted as $N_i$ where $i$ $\in$ [1, $n$]; $E_{ij}$ represents a set of $m$ edges that connects node $N_i$ and $N_j$, where $m$ denotes the number of operations. 
Any architecture candidate that can be found in the space has only one edge in $E_{ij}$. In other words, there is exactly one operation from $N_i$ to $N_j$ in any architecture candidate. In addition, an available architecture has at least one edge from its predecessor node.

\begin{table}
\makeatletter\def\@captype{table}\makeatother
\caption{Rank correlation analysis using Kendall's Tau~\cite{tau} for different split strategies.
}
\label{tab:split_strategy}
\centering
\resizebox{0.4\textwidth}{!}{

\begin{tabular}{@{}|c|c|c|c|@{}}
\toprule
\textbf{\#edges to split} & \textbf{\#split choices} & Mean  & Std.  \\ \midrule
1                       & 5                   & 0.653 & 0.012 \\ \midrule
2                       & 25                  & 0.696 & 0.016 \\ \midrule
3                       & 125                 & 0.752 & 0.018 \\ \bottomrule
\end{tabular}%
}
\end{table}

\paragraph{Split Procedure Analysis.}
Given a search space, e.g., one that was depicted in Figure~\ref{fig:supernet}, the full DAG contains $\frac{n(n-1)}{2}$ compound edges. Each compound edges has $m$ choices from the operations, resulting in a total of $m^{n(n-1)/2}$ architectures. Training all $m^{n(n-1)/2}$ architectures, as done by the vanilla NAS can provide accurate rank information but is time-consuming. 
To evaluate the impact of compound edge splitting on ranking architectures, we calculate the Kendall's Tau(rank correlation) for different split strategies on \nasbench (more details of this dataset in Section~\ref{nasbench201}).

Specifically, we reuse the supernet from \nasbench with 5 operation types. The supernet is a 4 nodes full DAG containing 6 independent compound edges, and each compound edge consists of the 5 parallel pre-defined operations. Hence, there are $C_{6}^{1}=6$, $C_{6}^{2}=15$, and $C_{6}^{3}=20$ choices to split 1, 2 and 3 edges. For example, there are 6 choices to split 1 compound edge, and each choice would generate $5^1$ sub-supernets because of 5 operations in a compound edge. We trained all 6, 15, and 20 edge choices with corresponding 1, 2, and 3 compound edges split to investigate the impact of split edge choices on rank correlation. 

Table~\ref{tab:split_strategy} shows the rank correlation when split with different numbers of compound edges. First, similar to what we have observed in Section~\ref{sec:bg_motivation}, increasing the number of split compound edges leads to a higher rank correlation. 
Second, given the same number of compound edges to split, the exact choice of which compound edge to split has a negligible impact on the rank correlation as indicated by the low standard deviation. 
Therefore, we can randomly choose which compound edge(s) to split and focus on how many compound edge(s) to split. In this work, we pre-define a training time budget $T$. If the total training time of supernet and all currently trained sub-supernets exceeds $T$, we will stop the split to avoid training more sub-supernets. Generally, $T$ is set to be twice of one-shot supernet training time.

\subsection{Transfer Learning}
\label{sub:transfer}

The number of sub-supernets grows exponentially with the number of split compound edges. Directly training all the resulting sub-supernets can be computationally intractable and also lose the benefit of one-shot NAS. In this section, we integrate the transfer learning technique to accelerate the training procedure of sub-supernets.

Similar to how an architecture candidate $\mathcal{A}$ inherits weights $\mathcal{W}_{\mathcal{A}}$ from the supernet weights $\mathcal{W}_{\mathcal{S}}$, we allow a sub-supernet $\mathcal{S}^{\Omega^{'}}$ to inherit weights from its parent sub-supernet. For example, in Figure~\ref{fig:motivation}(b), after training the supernet of $\Omega_{A}$, the supernet of $\Omega_{B}$ and $\Omega_{C}$ can inherit the weights from shared operations in supernet of $\Omega_{A}$ as initialization and then start training. Compared to training from scratch, each sub-supernet converges in only a few epochs with transfer learning.

\subsection{Integration with Gradient-based Algorithms}
\label{subsec:integration_gradient}

\paragraph{Gradient-based NAS Overview.}

Gradient-based algorithms work on a continuous search space, which can be converted from the DAG. Gradient-based algorithms treat NAS as a joint optimization problem where both the weight and architecture distribution parameters are optimized \emph{simultaneously} by training~\cite{DARTS}. In other words, gradient-based algorithms are designed for and used with the one-shot NAS.

To use gradient-based algorithms with our \name , 
we \emph{first} train the supernet until it converges. Then, we split the supernet $\mathcal{S}$ to several sub-supernets as described in Section~\ref{subsec:split_strategies} and initialize these sub-supernets with weights and architecture distribution parameters transferred from their parents. Next, we train these sub-supernets to converge and choose the sub-supernet $\mathcal{S}^{\Omega^{'}}$ with the lowest validation loss from all sub-supernets. Lastly, we pick the best architecture $\mathcal{A}^{*}$ from the $\mathcal{S}^{\Omega^{'}}$ based on the architecture distribution parameters.

\subsection{Integration with Search-based Algorithms}
\label{subsec:integration_search_based}

\paragraph{Search-based NAS Overview.} For search-based algorithms, a value function of candidate architecture is needed to guide the search. The value function can be non-differentiable and is often provided by either one-shot or vanilla NAS. For vanilla NAS, it is not strictly necessary to train these architectures to converge, and one can use early stopping to obtain an intermediate result. By starting with a few initial architectures, search-based algorithms sample the next architecture $\mathcal{A}$ from the search space based on previous sampled architectures and search algorithms until an architecture with satisfactory performance is found. 

To use search-based algorithms with our \name, we will first train a number of sub-supernets. Similar to what was described in Section~\ref{subsec:integration_gradient}, these converged sub-supernets will be used as the basis to evaluate the performance of sampled architectures. For example, if a sampled architecture $\mathcal{A}$ falls into sub-supernet $\mathcal{S}^{\Omega^{'}}$, we will evaluate its performance $f(\mathcal{S}^{\Omega^{'}}_{\mathcal{A}})$ by inheriting the weights $\mathcal{W}_{\mathcal{S}^{\Omega^{'}}}$. Once the search algorithms complete, we will pick the top $K$ architectures with the best performance empirically and train these architectures to converge and select the final architectures based on their performance.

\section{Experiments}
\label{experiment}

\begin{figure*}[h!]
\centering 
\includegraphics[width=1.92\columnwidth]{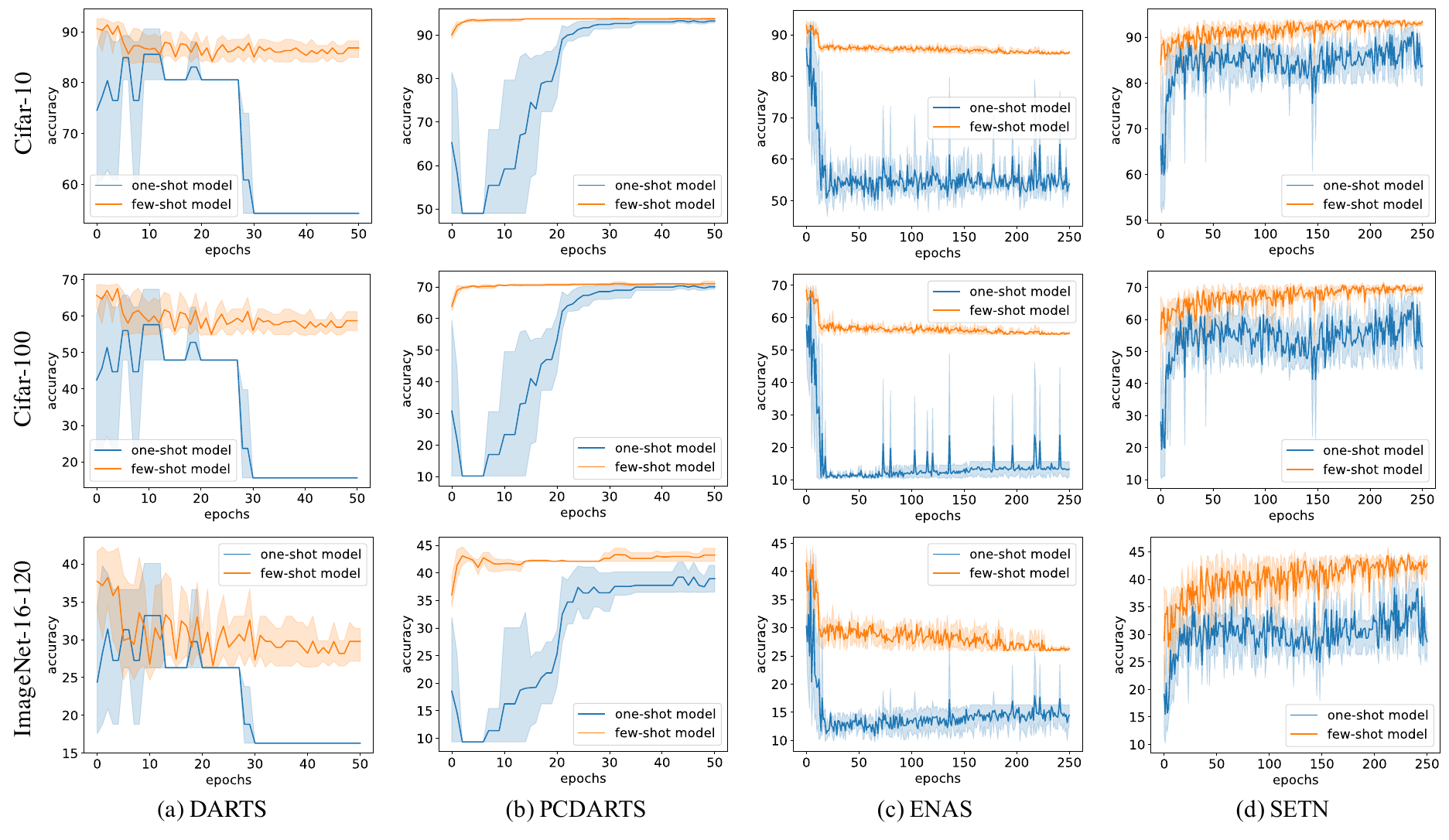}
\caption{Anytime accuracy comparison of state-of-the-art gradient-based algorithms on \name.  We ran each algorithm five times.} 
\label{fig:gradient-based-results}
\end{figure*}

To evaluate the performance of \name in reducing the approximation error associated with supernet and improving search efficiency of search algorithms, we conducted two types of evaluations. The first bases on an existing NAS dataset, and the second type compares the architectures found by using \name to state-of-the-art results in popular application domains.

We first evaluate the search performance of \name in different NAS algorithms. We use two metrics (search cost and accuracy) to evaluate search efficiency of DARTS, PCDARTS, ENAS, SETN, REA, REINFORCE, HB, BOHB, SMAC, and TPE~\cite{DARTS,pcdarts,enas,setn,rea,nasnet,hb,bohb,smac,tpe} by one-shot/few-shot models on \nasbench.
We also evaluate the search performance of \name with DARTS, PCDARTS, and ENAS on NasBench1-shot-1~\cite{nasbench1-shot-1}. Then we extend \name to different open domain search spaces and show that the found architectures significantly outperform the ones obtained by one-shot NAS. Our found architectures also reach state-of-the-arts results in CIFAR10, ImageNet, AutoGAN~\cite{autogan}, and Penn Treebank. 

\begin{figure*}[t]
\centering 
\includegraphics[width=1.92\columnwidth]{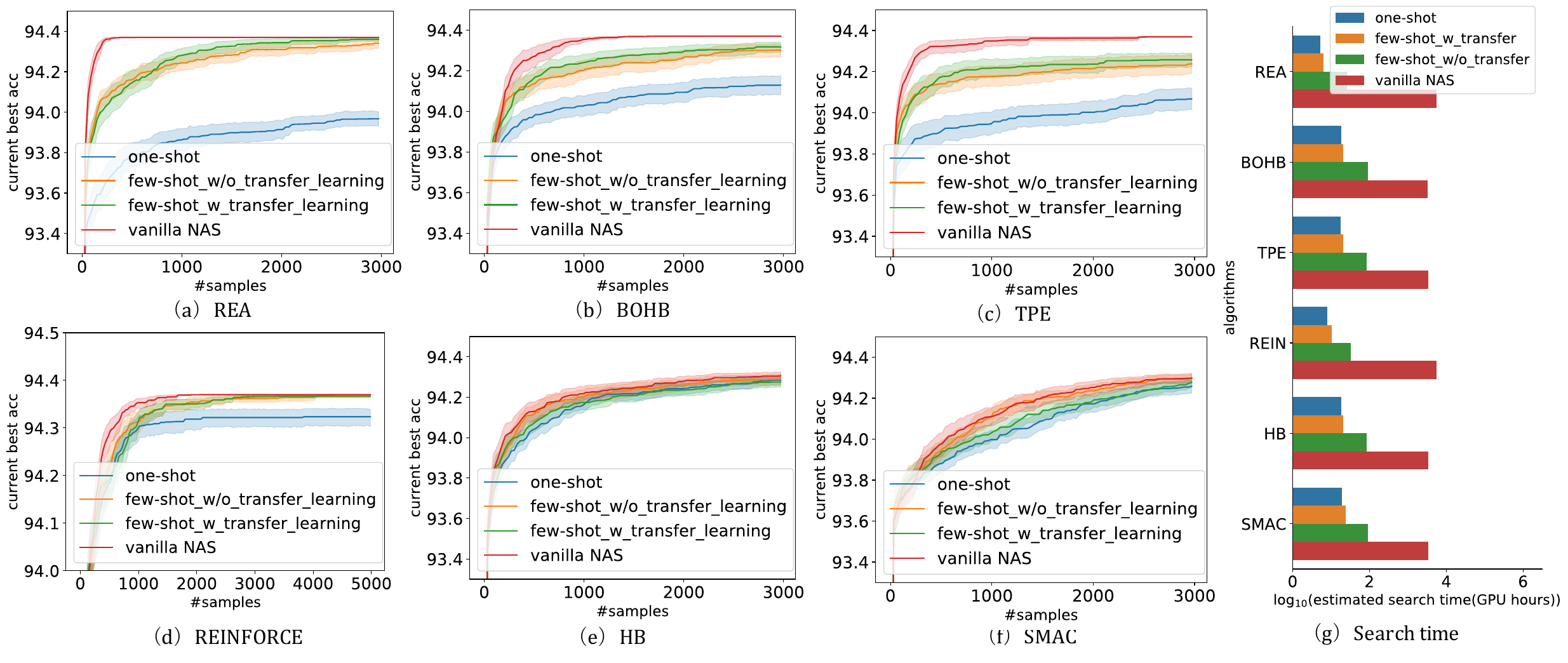}
\caption{Current best accuracy and search time comparison of popular search-based algorithms. All algorithms were ran for 50 times for one-shot, few-shot with/without transfer learning and vanilla NAS, respectively.}
\label{fig:search-based-models} 
\end{figure*}

\subsection{Evaluation on \nasbench}
\label{nasbench201}

We use \nasbench, a public architecture dataset, which provides a unified benchmark for up-to-date NAS algorithms~\cite{nasbench201}. 
\nasbench contains the information of \emph{all} 15625 architectures in the search space, making it possible to evaluate the efficiency of gradient-based search algorithms.
In contrast, prior datasets such as \nasbencho~\cite{nasbench101} do not provide all possible architectures information in their search space. For each architecture, \nasbench contains information such as size, training and test time, and accuracy on \cifara, \cifarb, and \imagenet. Consequently, NAS algorithms can leverage such information on each architecture without time-consuming training. 

\subsubsection{Gradient-based Algorithms}
\label{subsec:eval_gradient}

\paragraph{Methodology.} The supernet corresponding to \nasbench has four nodes and five operations. Based on the split method described in Section~\ref{subsec:split_strategies}, we split one compound edge (i.e., parallel operations) in search space and obtain five sub-supernets. For this experiment, we train sub-supernets by skipping the transfer learning described in Section~\ref{sub:transfer}. Doing so allows us to compare the anytime performance between one-shot and \name, by keeping the same training epochs. Due to the limit of the computing resource, we only split one compound edge in the search space since training more sub-supernets exponentially increases the time cost without transfer learning. 

We chose a number of recently proposed gradient-based search algorithms, including DARTS, ENAS, PCDARTS, and SETN~\cite{DARTS, enas, pcdarts, setn} for evaluating the search performance of \name. We used two metrics to compare the search performance between one-shot and \name: \1 \emph{test accuracy} which is obtained by evaluating the final architecture found by a NAS algorithm;
and \2 \emph{search time} that includes the supernet training and validation time.

\paragraph{Result Analysis.} Figure~\ref{fig:gradient-based-results} shows the anytime test accuracy of searched models. In the case of training on \cifara, when using one-shot NAS, both the architectures found by DARTS and ENAS can easily trap in a bad performance region with the search progress, an observation that is consistent with the original paper~\cite{nasbench201}. The potential reason is one-shot NAS has an inaccurate performance prediction. In contrast, \name could obtain higher-quality searched models than one-shot NAS. The good performance of \name is likely due to using multiple supernets, allowing us to have a more accurate performance estimation to guide the search. Additionally, in the case of PCDARTS and SETN, even though one-shot NAS was able to eventually find a good architecture, i.e., with more than 90\% accuracy, it took 10X more search epochs than our \name. In short, we show that by using \name, gradient-based algorithms can have more effective and efficient search in terms of found architectures and the number of search epochs.

\subsubsection{Search-based Algorithms}

\paragraph{Methodology.}
As described in Section~\ref{subsec:split_strategies}, we define the search time budget of \name to be twice as that of one-shot NAS. Therefore, we only split the supernet by the one compound edge between the first node $N_0$ and second node $N_1$(i.e $E_{01}$ in Figure~\ref{fig:supernet}) to five sub-supernets. 
Note that, in this experiment, we trained all sub-supernets with transfer learning described in Section~\ref{sub:transfer}.  
We chose six different search-based algorithms including REA, REINFORCE, BOHB , HB, SMAC, and TPE~\cite{rea,nasnet,hb,bohb,smac,tpe} and ran each search algorithm 50 times. 
We evaluated the effectiveness of \name by following the procedure described in Section~\ref{subsec:integration_search_based}.
We used two metrics to evaluate the performance of search-based algorithms. The first metric is \emph{$i^{th}$ best accuracy} which denotes the best test accuracy after searching all $i$ architectures. This metric helps to quantify the search efficiency, as a good search algorithm is expected to find an architecture with higher test accuracy with fewer samples. 
The second metric is \emph{total search time} which includes the training time of supernet (for one-shot NAS) and sub-supernets (for \name) and the time for finding the satisfiable architecture(s).

\begin{table}
\makeatletter\def\@captype{table}\makeatother\caption{Rank correlation analysis using Kendall's Tau on \nasbench with different methods. Our \name has a great improvement of Kendall's Tau compared with other baselines. Note that more supernet splits enhance the rank correlation but the overheads are increased accordingly.}
\label{tab:rank}
\resizebox{0.48\textwidth}{!}{
\begin{tabular}{|c|c|c|}
\hline
Method                           & Kendall’s Tau  & \multicolumn{1}{l|}{Cost(Hours)} \\ \hline
Random                           & 0.0022         & 0                                \\ \hline
EN$^2$AS~\cite{en2as}                             & 0.378          & N/A                              \\
One-shot                         & 0.5436         & 6.8                              \\
Angle~\cite{angle-based}                            & 0.5748         & N/A                              \\ \hline
\textbf{Few-shot(5-supernets)}   & \textbf{0.653} & 10.1                              \\
\textbf{Few-shot(25-supernets)}  & \textbf{0.696} & 18.6                              \\
\textbf{Few-shot(125-supernets)} & \textbf{0.752} & 31.8                             \\ \hline
\end{tabular}%
}
\end{table}

\paragraph{Result Analysis.} Figure~\ref{fig:search-based-models}(a)-(f) compare the best accuracy after searching a certain number of architectures. We \emph{first} observe that \name was able to find the best architecture in the search space with about 3500 samples when using REA, and 3000 samples when using REINFORCE. Second, we see that with REA, REINFORCE, BOHB, and TPE, \name significantly improved the search efficiency over one-shot NAS. Lastly, with HB and SMAC, both NAS algorithms achieved slightly better search efficiency with \name compared to using one-shot NAS.

Figure~\ref{fig:search-based-models}(g) compares the total search time. All search-based algorithms took three to four orders of magnitude GPU hours when using vanilla NAS, compared to both one-shot NAS and our \name. 
Further, \name took similar search time compared to one-shot NAS, both completing the search within 24 hours. 
We also observe that, by using transfer learning, \name saved a significant amount of search time compared to \name without transfer learning.
Such time saving with transfer learning was achieved with negligible search performance variation in terms of current best accuracy. 
Together, these results suggest that transfer learning is a valuable addition to our \name design. 

Finally, we compare the rank correlation (the higher, the better) in Table~\ref{tab:rank}. 
We show that \name, with the different number of sub-supernets, achieved better rank correlation than prior work. 
The good correlation achieved by \name indicates its effectiveness in finding higher accuracy architecture.

\begin{figure}[t]

\centering 
\subfloat[][DARTS]{\includegraphics[height=1.12in]{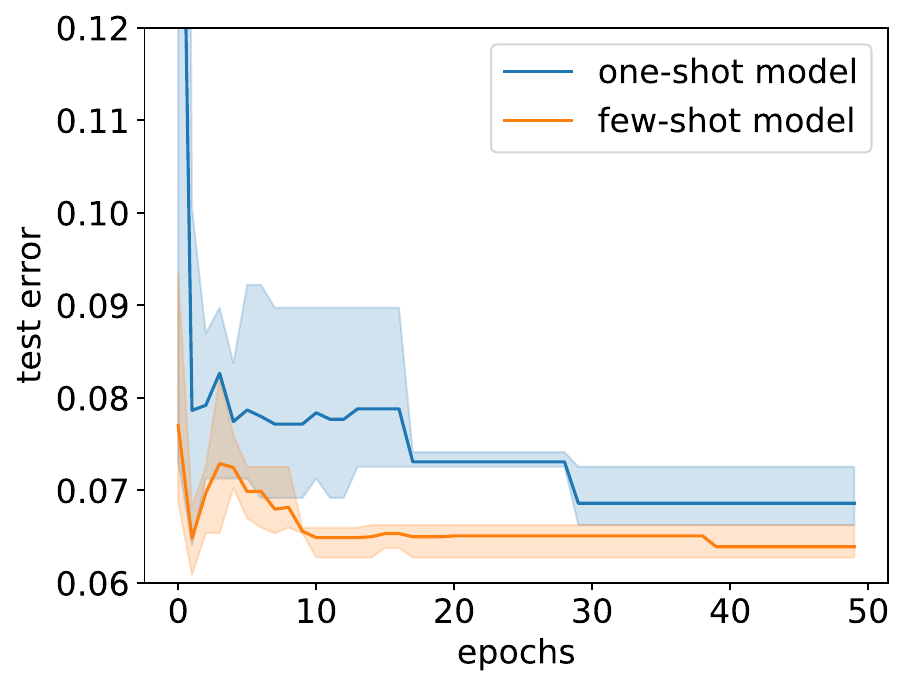}\label{darts}} \quad
\subfloat[][PCDARTS]{\includegraphics[height=1.12in]{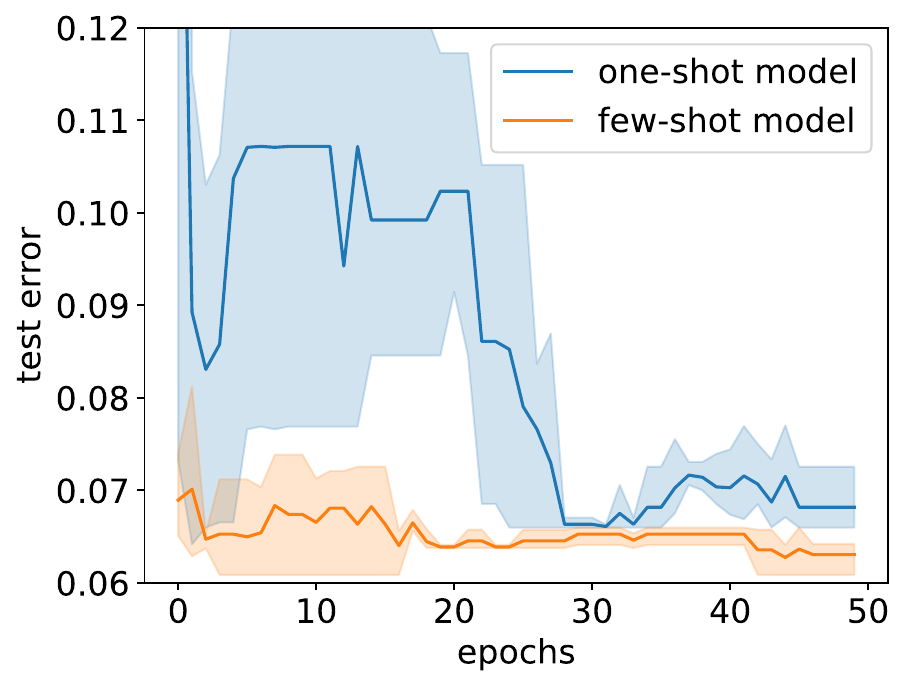}\label{pcdarts}}  \quad
\subfloat[][ENAS]{\includegraphics[height=1.12in]{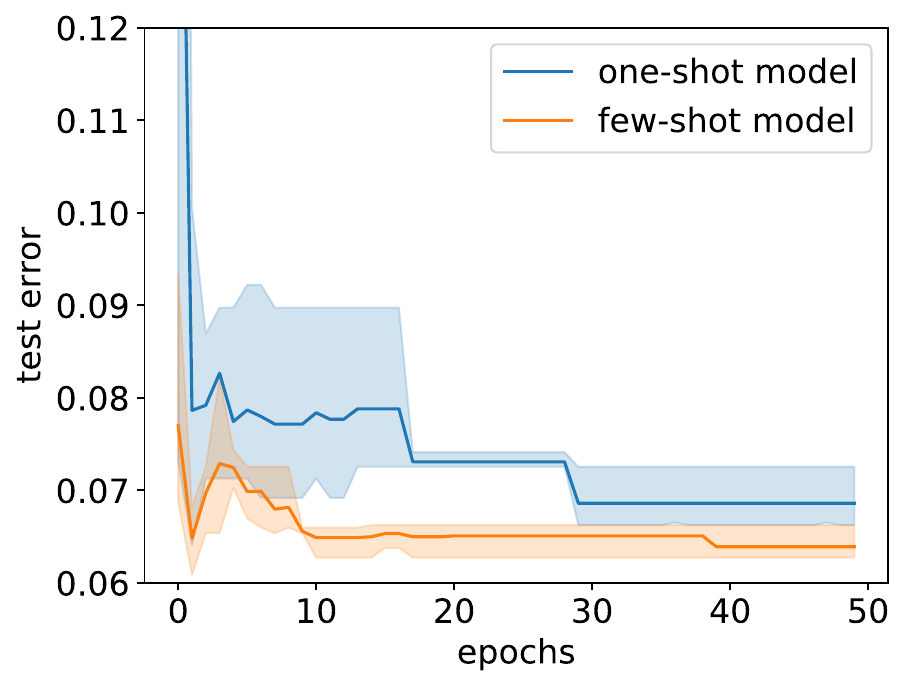}\label{enas}}  \quad
\subfloat[][RANDOM]{\includegraphics[height=1.12in]{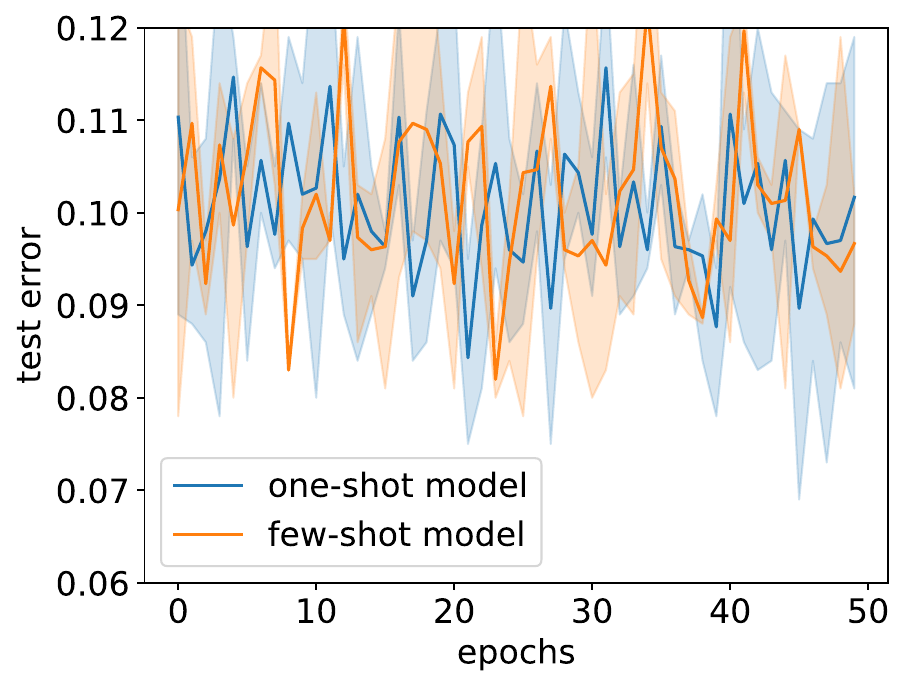}\label{random}}  \quad
\caption{Search results on Nasbench1-shot-1. We ran each search algorithm 3 times.}
\label{1shot1}
\end{figure}

\subsection{Evaluation on NasBench1-shot-1}

We evaluated our \name on NasBench1-shot-1~\cite{nasbench1-shot-1}, which is a public neural architecture dataset similar to \nasbench. Instead of creating a new search space like \nasbench, NasBench1-shot-1 supports one-shot NAS algorithms by directly leveraging \nasbencho~\cite{nasbench101}.
In this experiment, we used three sub-supernets in our \name and kept all other settings the same as Section~\ref{nasbench201}. 

\paragraph{Result Analysis.} Figure~\ref{1shot1} shows search results on NasBench1-Shot-1. Our \name improved the search efficiency both in terms of the test accuracy of found architectures and search time. Specifically,
with DARTS, PCDARTS, and ENAS, our \name can quickly find good architectures, i.e., finding architectures with test error less than 0.07 in about 10 epochs; while one-shot PCDARTS requires near 30 epochs to find an architecture with similar performance. The final searched architectures by \name also outperform the counterparts by one-shot NAS.  

\begin{table}
 \makeatletter\def\@captype{table}\makeatother\caption{Applying \name on existing NAS methods on CIFAR-10 using the NASNet search space. Our results demonstrate that 1) \name consistently improves the final accuracy of various one-shot-based NAS methods under the same setup. Please note we only extend one-shot based DARTS, REA, and LaNAS by replacing the single supernet with 7 supernets in their public release;
 2) after integrating with multiple supernets, few-shot DARTS achieves SOTA 98.72\% top-1 accuracy on CIFAR-10 using the cutout~\cite{cutout} and auto-augmentation~\cite{autoaugment}. Without auto-augmentation, few-shot DARTS-Small still consistently outperforms existing models that have similar parameters.}
 
\label{acc-comps-cifar}
\resizebox{0.46\textwidth}{!}{
\begin{tabular}{@{}ccccc@{}}
\toprule
\textbf{Method}                                      & \textbf{Data Augmentation}              & \textbf{\#Params}          & \textbf{Err}                                & \textbf{GPU days}         \\ \midrule
\multicolumn{1}{|c|}{NASNet-A~\cite{nasnet}}                       & \multicolumn{1}{c|}{cutout} & \multicolumn{1}{c|}{3.3M}  & \multicolumn{1}{c|}{2.65}                   & \multicolumn{1}{c|}{2000} \\
\multicolumn{1}{|c|}{AmoebaNet-B-small~\cite{rea}}                    & \multicolumn{1}{c|}{cutout} & \multicolumn{1}{c|}{2.8M}  & \multicolumn{1}{c|}{2.50$\pm$0.05}          & \multicolumn{1}{c|}{3150} \\
\multicolumn{1}{|c|}{AmoebaNet-B-large~\cite{rea}}                    & \multicolumn{1}{c|}{cutout} & \multicolumn{1}{c|}{34.9M}  & \multicolumn{1}{c|}{2.13$\pm$0.04}          & \multicolumn{1}{c|}{3150} \\
\multicolumn{1}{|c|}{AlphaX~\cite{alphax}}                     & \multicolumn{1}{c|}{cutout} & \multicolumn{1}{c|}{2.83M}  & \multicolumn{1}{c|}{2.54$\pm$0.06}                   & \multicolumn{1}{c|}{1000}  \\
\multicolumn{1}{|c|}{NAO~\cite{nao}}                    & \multicolumn{1}{c|}{cutout} & \multicolumn{1}{c|}{3.2M}  & \multicolumn{1}{c|}{3.14$\pm$0.09}          & \multicolumn{1}{c|}{225} \\
\midrule
\multicolumn{1}{|c|}{DARTS~\cite{DARTS}}                          & \multicolumn{1}{c|}{cutout} & \multicolumn{1}{c|}{3.3M}  & \multicolumn{1}{c|}{2.76$\pm$0.09}          & \multicolumn{1}{c|}{1}    \\
\multicolumn{1}{|c|}{P-DARTS~\cite{pdarts}}                          & \multicolumn{1}{c|}{cutout} & \multicolumn{1}{c|}{3.4M}  & \multicolumn{1}{c|}{2.5}          & \multicolumn{1}{c|}{0.3}    \\
\multicolumn{1}{|c|}{PC-DARTS~\cite{pcdarts}}                          & \multicolumn{1}{c|}{cutout} & \multicolumn{1}{c|}{3.6M}  & \multicolumn{1}{c|}{2.57$\pm$0.07}          & \multicolumn{1}{c|}{0.3}    \\
\multicolumn{1}{|c|}{Fair-DARTS~\cite{fairdarts}}                          & \multicolumn{1}{c|}{cutout} & \multicolumn{1}{c|}{3.32M}  & \multicolumn{1}{c|}{2.54$\pm$0.05}          & \multicolumn{1}{c|}{3}    \\
\multicolumn{1}{|c|}{BayeNAS~\cite{bayesnas}}                          & \multicolumn{1}{c|}{cutout} & \multicolumn{1}{c|}{3.4M}  & \multicolumn{1}{c|}{2.81$\pm$0.04}          & \multicolumn{1}{c|}{0.2}    \\
\multicolumn{1}{|c|}{CNAS~\cite{cnas}}                          & \multicolumn{1}{c|}{cutout} & \multicolumn{1}{c|}{3.7M}  & \multicolumn{1}{c|}{2.60$\pm$0.06}          & \multicolumn{1}{c|}{0.3}    \\
\multicolumn{1}{|c|}{MergeNAS~\cite{mergenas}}                          & \multicolumn{1}{c|}{cutout} & \multicolumn{1}{c|}{2.9M}  & \multicolumn{1}{c|}{2.68$\pm$0.01}          & \multicolumn{1}{c|}{0.6}    \\
\multicolumn{1}{|c|}{ASNG-NAS~\cite{asng-nas}}                          & \multicolumn{1}{c|}{cutout} & \multicolumn{1}{c|}{3.32M}  & \multicolumn{1}{c|}{2.54$\pm$0.05}          & \multicolumn{1}{c|}{0.11}    \\
\multicolumn{1}{|c|}{XNAS~\cite{xnas}}                     & \multicolumn{1}{c|}{cutout + autoaug} & \multicolumn{1}{c|}{3.7M}  & \multicolumn{1}{c|}{1.81}                   & \multicolumn{1}{c|}{0.3}  \\
\multicolumn{1}{|c|}{one-shot REA}                  & \multicolumn{1}{c|}{cutout + autoaug} & \multicolumn{1}{c|}{3.5M}  & \multicolumn{1}{c|}{2.02$\pm$0.03}          & \multicolumn{1}{c|}{0.75} \\
\multicolumn{1}{|c|}{one-shot LaNas~\cite{lanas}}                & \multicolumn{1}{c|}{cutout + autoaug} & \multicolumn{1}{c|}{3.6M}  & \multicolumn{1}{c|}{1.68$\pm$0.06}                   & \multicolumn{1}{c|}{3}  \\
\midrule
\multicolumn{1}{|c|}{\textbf{few-shot DARTS-Small}} & \multicolumn{1}{c|}{cutout} & \multicolumn{1}{c|}{3.8M}  & \multicolumn{1}{c|}{\textbf{2.31$\pm$0.08}} & \multicolumn{1}{c|}{1.35}  \\
\multicolumn{1}{|c|}{\textbf{few-shot DARTS-Large}} & \multicolumn{1}{c|}{cutout} & \multicolumn{1}{c|}{45.5M} & \multicolumn{1}{c|}{\textbf{1.92$\pm$0.08}} & \multicolumn{1}{c|}{1.35}  \\
\multicolumn{1}{|c|}{\textbf{few-shot DARTS-Small}} & \multicolumn{1}{c|}{cutout + autoaug} & \multicolumn{1}{c|}{3.8M}  & \multicolumn{1}{c|}{\textbf{1.70$\pm$0.08}} & \multicolumn{1}{c|}{1.35}  \\
\multicolumn{1}{|c|}{\textbf{few-shot DARTS-Large}} & \multicolumn{1}{c|}{cutout + autoaug} & \multicolumn{1}{c|}{45.5M} & \multicolumn{1}{c|}{\textbf{1.28$\pm$0.08}} & \multicolumn{1}{c|}{1.35}  \\
\multicolumn{1}{|c|}{\textbf{few-shot REA}}         & \multicolumn{1}{c|}{cutout + autoaug} & \multicolumn{1}{c|}{3.7M}  & \multicolumn{1}{c|}{\textbf{1.81$\pm$0.05}} & \multicolumn{1}{c|}{0.87} \\
\multicolumn{1}{|c|}{\textbf{few-shot LaNas}}       & \multicolumn{1}{c|}{cutout + autoaug} & \multicolumn{1}{c|}{3.2M}  & \multicolumn{1}{c|}{\textbf{1.58$\pm$0.04}} & \multicolumn{1}{c|}{3.8}  \\ 
\bottomrule
\end{tabular}%

}
\end{table}

\subsection{Deep Learning Applications}

\paragraph{\cifara in Practice.} We chose three state-of-the-art NAS algorithms, one gradient-based algorithm(DARTs) and two search-based algorithms, including regularized evolution (REA) and LaNas~\cite{DARTS, rea,lanas}, for evaluating the effectiveness of \name.  
We used the same search space based on the original DARTS, REA, and LaNas.

Table~\ref{acc-comps-cifar} compares the search performance achieved by \name to recently proposed NAS algorithms. 
We observe that \name substantially improves the test accuracy across DARTS, REA, and LaNas. Specifically, we see that \name with DARTS outperformed searching directly with one-shot (original) DARTS by 0.43 lower error. Further, the architecture found with \name also outperforms the state-of-the-art results on \cifara. Additionally, \name only incurred a 35\% search time increase. Similarly, \name also improved the search efficiency of REA upon one-shot NAS, finding an architecture with 0.21 lower error with only 16.7\% more search time. For LaNas, \name decreases the test error from 1.68 to 1.58 with only 26.7\% extra search time. All of our few-shot searched architectures outperform the architectures of other methods in the table with the same training setup. In short, \name improved the efficiency of existing state-of-the-art search algorithms.

\paragraph{Neural Architecture Search on ImageNet.} We selected two state-of-the-art NAS algorithms that were designed for the ImageNet, including ProxylessNAS and Once-for-All NAS (OFA)~\cite{proxylessnas, ofa}.
We used the same training setup for \name as ProxylessNAS and OFA.

Table~\ref{acc-comps-imagenet} compares the search performance. 
First, we can see that our few-shot NAS significantly improved the accuracy and kept similar FLOPs numbers on both two NAS algorithms with the one-shot NAS. 
Second, our few-shot OFA achieves the best top-1 accuracy (with the same scale of FLOPs numbers) compared to architectures found with all other search methods.

\paragraph{Comparison to AUTOGAN~\cite{autogan}.} AUTOGAN was proposed to search for a special architecture called GAN, which consists of two competing networks. The networks, a generator, and a discriminator play a min-max two-player game against each other. We followed the same setup described in the AUTOGAN paper. We used Inception score (IS) (higher is better) and Frchet Inception Distance (FID)~\cite{fid} (lower is better) to evaluate the performance of GAN. Table~\ref{AutoGan_table} compares the top three performing GANs found by both original AUTOGAN and using our \name. We observe that by using \name, the inception score of the best architecture was improved from 8.55 to 8.63, and the FID was reduced from 12.42 to 10.73. Additionally, the top two architectures found using \name had very close performance, one with the lowest inception score and the other with the lowest FID. In short, all three architectures found by \name had better inception score and FID than state-of-the-art results.

\begin{table}
 \makeatletter\def\@captype{table}\makeatother\caption{ Applying \name on existing NAS methods on ImageNet using the EfficientNet search space. Being consistent with the results on CIFAR-10 in Table.~\ref{acc-comps-cifar}, the final accuracy from few-shot OFA and ProxylessNAS also outperforms their original one-shot version under the same setting, except for replacing the single supernet with 5 supernets. Particularly, Few-shot OFA-Large achieves SoTA 80.5\% top1 accuracy at 600M FLOPS.}
\label{acc-comps-imagenet}
\resizebox{0.46\textwidth}{!}{
\begin{tabular}{cccccc}
\hline
\textbf{Method}                                       & \textbf{Space}              & \textbf{\#Params}          & \textbf{\#FLOPs}          & \textbf{Top 1 Acc(\%)}         & \textbf{GPU hours}      \\ \hline

\multicolumn{1}{|c|}{AutoSlim~\cite{autoslim}}                    & \multicolumn{1}{c|}{Mobile} & \multicolumn{1}{c|}{5.7M} & \multicolumn{1}{c|}{305M} & \multicolumn{1}{c|}{74.2}     & \multicolumn{1}{c|}{N/A}     \\
\multicolumn{1}{|c|}{MobileNetV3-Large~\cite{mobilenetv3}}                    & \multicolumn{1}{c|}{Mobile} & \multicolumn{1}{c|}{5.4M} & \multicolumn{1}{c|}{219M} & \multicolumn{1}{c|}{74.7}    & \multicolumn{1}{c|}{N/A}       \\
\multicolumn{1}{|c|}{MnasNet-A2~\cite{mnasnet}}                    & \multicolumn{1}{c|}{Mobile} & \multicolumn{1}{c|}{4.8M} & \multicolumn{1}{c|}{340M} & \multicolumn{1}{c|}{75.6}   & \multicolumn{1}{c|}{N/A}        \\
\multicolumn{1}{|c|}{FBNetV2-L1~\cite{fbnetv2}}                    & \multicolumn{1}{c|}{Mobile} & \multicolumn{1}{c|}{N/A} & \multicolumn{1}{c|}{325M} & \multicolumn{1}{c|}{77.2}     & \multicolumn{1}{c|}{600}      \\
\multicolumn{1}{|c|}{EfficientNetB0~\cite{efficientnet}}                    & \multicolumn{1}{c|}{Mobile} & \multicolumn{1}{c|}{5.3M} & \multicolumn{1}{c|}{390M} & \multicolumn{1}{c|}{77.3}     & \multicolumn{1}{c|}{N/A}      \\

\multicolumn{1}{|c|}{AtomNAS~\cite{atomnas}}                    & \multicolumn{1}{c|}{Mobile} & \multicolumn{1}{c|}{5.9M} & \multicolumn{1}{c|}{363M} & \multicolumn{1}{c|}{77.6}    & \multicolumn{1}{c|}{N/A}       \\
\multicolumn{1}{|c|}{\textbf{few-shot OFA\_Net-Small}}   & \multicolumn{1}{c|}{Mobile} & \multicolumn{1}{c|}{5.6M}   & \multicolumn{1}{c|}{238M} & \multicolumn{1}{c|}{\textbf{77.50}} & \multicolumn{1}{c|}{68} \\\hline
\multicolumn{1}{|c|}{MobileNetV2~\cite{mobilenetv2}}                    & \multicolumn{1}{c|}{Mobile} & \multicolumn{1}{c|}{6.9M} & \multicolumn{1}{c|}{585M} & \multicolumn{1}{c|}{74.7}    & \multicolumn{1}{c|}{N/A}       \\
\multicolumn{1}{|c|}{ShuffleNet-V2~\cite{shufflenetv2}}                    & \multicolumn{1}{c|}{Mobile} & \multicolumn{1}{c|}{N/A} & \multicolumn{1}{c|}{590M} & \multicolumn{1}{c|}{74.9}    & \multicolumn{1}{c|}{N/A}       \\
\multicolumn{1}{|c|}{ProxylessNAS~\cite{proxylessnas}}                    & \multicolumn{1}{c|}{Mobile} & \multicolumn{1}{c|}{7.12M} & \multicolumn{1}{c|}{465M} & \multicolumn{1}{c|}{75.1}     & \multicolumn{1}{c|}{200}      \\
\multicolumn{1}{|c|}{ChamNet~\cite{chamnet}}                    & \multicolumn{1}{c|}{Mobile} & \multicolumn{1}{c|}{N/A} & \multicolumn{1}{c|}{553M} & \multicolumn{1}{c|}{75.4}    & \multicolumn{1}{c|}{N/A}      \\
\multicolumn{1}{|c|}{RegNet~\cite{regnet}}                    & \multicolumn{1}{c|}{Mobile} &
\multicolumn{1}{c|}{6.1M} & \multicolumn{1}{c|}{600M} & \multicolumn{1}{c|}{75.5}      & \multicolumn{1}{c|}{N/A}     \\

\multicolumn{1}{|c|}{OFA\_Net~\cite{ofa}}                        & \multicolumn{1}{c|}{Mobile} & \multicolumn{1}{c|}{9.1M}   & \multicolumn{1}{c|}{595M} & \multicolumn{1}{c|}{80.0}     & \multicolumn{1}{c|}{40}      \\ \hline
\multicolumn{1}{|c|}{\textbf{few-shot ProxylessNAS}} & \multicolumn{1}{c|}{Mobile} & \multicolumn{1}{c|}{4.87M} & \multicolumn{1}{c|}{521M} & \multicolumn{1}{c|}{\textbf{75.91}} & \multicolumn{1}{c|}{280} \\
\multicolumn{1}{|c|}{\textbf{few-shot OFA\_Net-Large}}   & \multicolumn{1}{c|}{Mobile} & \multicolumn{1}{c|}{9.2M}   & \multicolumn{1}{c|}{600M} & \multicolumn{1}{c|}{\textbf{80.50}} & \multicolumn{1}{c|}{68} \\\hline
\end{tabular}%
}
\end{table}

\begin{table}
        \makeatletter\def\@captype{table}\makeatother\caption{Applying \name to AutoGAN by only replacing the supernet with 3 supernets in their public release. Few-shot AutoGAN demonstrates up to 20\% better performance than the original one-shot AutoGAN.}
        \label{AutoGan_table}
        \resizebox{0.455\textwidth}{!}{
                 \begin{tabular}{@{}|c|c|c|@{}}
\toprule
\textbf{Method}                       & \textbf{Inception Score} & \textbf{FID Score}  \\ \midrule
ProbGAN\cite{probgan}                               & 7.75$\pm$.14             & 24.60         \\
SN-GAN\cite{sngan}                              & 8.22$\pm$.05             & 21.70$\pm$.01 \\
MGAN\cite{mgan}                                  & 8.33$\pm$.12             & 26.7          \\
Improving MMD GAN\cite{mmdgan}                     & 8.29                     & 16.21         \\ \midrule
AutoGAN-top1\cite{autogan}                          & 8.55$\pm$.10             & 12.42         \\
AutoGAN-top2                          & 8.42$\pm$.06             & 13.67         \\
AutoGAN-top3                          & 8.41$\pm$.12             & 13.87         \\ \midrule
\textbf{few-shot AutoGAN-top1} & 8.60$\pm$.10             & \textbf{10.73$\pm$.10} \\
\textbf{few-shot AutoGAN-top2} & \textbf{8.63$\pm$.09}             & 10.89$\pm$.20 \\
\textbf{few-shot AutoGAN top3} & 8.52$\pm$.08             & 12.20         \\ \bottomrule
\end{tabular}

}
\end{table}

\paragraph{PENN TREEBANK in Practice~\cite{ptb}.}
Lastly, we evaluate \name on Penn Treebank (PTB), a widely-studied benchmark for language models. We used the same search space and training setup as the original DARTS to search RNN on PTB. By using \name, we achieved the state-of-the-art test Perplexity of 54.89 with an overall cost of 1.56 GPU days. In comparison, the original DARTS used 1 GPU day to find an architecture with worse performance (55.7 test Perplexity).

\section{Related Works}

Weight-sharing supernet was first proposed as a way to reduce the computational cost of NAS~\cite{enas}. Centering around supernet, a number of NAS algorithms including gradient-based~\cite{DARTS,pcdarts,setn} and search-based~\cite{BG_understanding,FairNAS,uniform_random_train} were proposed. The search efficiency of these algorithms is dependent on the ability of the supernet to approximate architecture performance.

To improve the supernet approximation accuracy, Bender et al.~\cite{BG_understanding} proposed a path dropout strategy that randomly drops out weights of the supernet during training. This approach improves the correlation between one-shot NAS and individual architecture accuracy by reducing weight co-adaptation. In a similar vein, Guo et al.~\cite{uniform_random_train} proposed a single-path one-shot training by only activating the weights from one randomly picked architecture in forward and backward propagation. Additionally, Yu et al.~\cite{how_to_train} found that training setup greatly impacts supernet performance and identified useful parameters and hyper-parameters. Lastly, an angle-based approach~\cite{angle1,angle2,angle3} was proposed to improve the supernet approximation accuracy for individual architecture~\cite{angle-based} and was shown to improve the architecture rank correlation. 
However, our \name achieved better rank correlation than this angle-based approach (see Table~\ref{tab:rank}). Our work focuses on reducing the supernet approximation error by dividing the supernet into a few sub-supernets to eliminate the co-adaption among supernet operations. As such, our work is complementary and can be integrated into the aforementioned work.

\section{Conclusion}

In this work, we proposed a novel approach, \name, for fast evaluation of candidate network architectures in Neural Architecture Search. \Name uses a few sub-supernets to efficiently and accurately evaluate the performance of candidate architectures, balancing the fast but low-quality evaluation in one-shot NAS with the expensive yet accurate evaluation in vanilla NAS. As a general evaluator, \Name can be integrated with both gradient-based and search-based optimization of neural architecture. Our extensive evaluations on a recent NAS benchmark \nasbench, NasBench1-shot-1 and deep learning applications demonstrated that \name significantly improved search performance of all popular one-shot methods with negligible search time overhead. Furthermore, the final results from \name also outperform previously published results by DARTs, ProxylessNas, Once-for-all NAS, and AUTO-GAN. 

\section*{Acknowledgements}
This work was supported in part by National Science Foundation Grants \#1755659 and \#1815619.

\bibliographystyle{icml2021}
\bibliography{fewshot}


\clearpage
\appendix

\section{Additional Notations}

We use two additional notations for pseudocode description: 
\1 $\mathcal{S}_{id}$ denotes a set of sub-supernets that is split by $id$ numbers of edges. 
\2 $\mathcal{S}_{id}^{n}$ denotes the $n^{th}$ sub-supernet in $\mathcal{S}_{id}$.

\section{End-to-end Pipeline Pseudocode}

Below we list the pseudocode for the end-to-end split and training pipeline in Algorithm~\ref{Pipeline}, the pseudocode for random split the one-shot model into sub-supernets in Algorithm~\ref{algo:split}, and the pseudocode for training (sub-)supernets in Algorithm~\ref{algo:train}.

\begin{algorithm}[H]
  \caption{(Sub)-supernets split and training}
  \label{Pipeline}
    \begin{algorithmic}[1]
      \STATE $\mathcal{S}_{0} = \{\mathcal{S}\}$ 
      \STATE define global $T \leftarrow TIME\_{BUDGET}$
      \STATE Train($\mathcal{S}$, NONE)
      \STATE $\mathcal{S}_{0} \leftarrow$ $\mathcal{S}$
      \STATE $id \leftarrow$ $0$
      \WHILE{total time $ < T$}
      \STATE $j \leftarrow$ $random(0, \#N)$
      \STATE $i \leftarrow$ $random(0, j)$
      \STATE $\mathcal{S}_{id+1} \leftarrow$ RandomSplit$(\mathcal{S}_{id}, E_{ij})$
      \FOR{$n = 1 \rightarrow sizeof(\mathcal{S}_{id+1})$}
        \STATE Train($\mathcal{S}_{id+1}^{n}$, $\mathcal{S}_{id}^{'}$)
      \ENDFOR
      \STATE $id \leftarrow$ $id + 1$
      \ENDWHILE

    \end{algorithmic}
\end{algorithm}

\begin{algorithm}[H]
  \caption{RandomSplit($\mathcal{S}_{id}, E_{ij}$)}

  \label{algo:split}
    \begin{algorithmic}[1]
    \STATE \quad \quad $\mathcal{S}_{new}$ $\leftarrow$ split $\mathcal{S}_{id}$ to $m$ sub-supernets given $m$ operations
    \STATE \quad \quad return $\mathcal{S}_{new}$
   
    \end{algorithmic}
\end{algorithm}

\begin{algorithm}[H]
  \caption{Train($s$, $parent$)}
  \label{algo:train}
    \begin{algorithmic}[1]
      \STATE  \textbf{if} $parent$ IS NOT NONE  \textbf{then}
      \STATE  \quad $W_{s} \leftarrow W_{parent}$
      \STATE  \textbf{end if} 
      \STATE  \textbf{While} $s$ NOT CONVERGE  \textbf{do}
      \STATE  \quad forward($s$)
      \STATE  \quad backward($s$)
      \STATE  \textbf{end While}
      
    \end{algorithmic}
\end{algorithm}

\section{Experiment Setup for Section~\ref{sec:bg_motivation}}

Each architecture was trained for 150 epochs with a batch size of 128. The initial channel is 16. We used the SGD optimizer with an initial learning rate of 0.025, followed by a cosine learning rate schedule through the training. We set the momentum rate to 0.9 and a weight decay of $3\times10^{-4}$. The training setup of supernet and sub-supernets is consistent with architecture candidates. These experiments ran on 50 P100 GPUs.  

\section{Experiment Setup for Section~\ref{experiment}}

\paragraph{(Sub-)supernet Training Setup for \nasbench.} Each architecture was trained for 200 epochs with 256 batch size. The initial channel is 16. We used the SGD optimizer with an initial learning rate of 0.1, followed by a cosine learning rate schedule through the training. The momentum rate was set to 0.9. We used a weight decay of $5 \times 10^{-4}$ and a norm gradient clipped at 5. The cutout technique was not used in training. The supernet training setup is consistent with architecture candidates. For supernet training, we changed the initial learning rate to 0.025 and total epochs to 300. The batch size is 128, and the weight decay was set to $1 \times 10^{-4}$. Each sub-supernet approximately took 40-50 epochs to converge after transfer learning. For each NAS algorithm, we used the same setup as described in the \nasbench~\cite{nasbench201}. We used 6 P100 GPUs to train the supernet and 5 sub-supernets.

\paragraph{Search Setup for DARTS on \cifara.} We used the same search space and training setup as described in the original DARTS paper~\cite{DARTS}. Specifically, the available operations in the search space include 3 x 3 and 5 x 5 separable convolutions, 3 x 3 and 5 x 5 dilated separable convolutions, 3 x 3 max pooling, 3 x 3 average pooling, identity, and zero. We trained 8 cells using DARTS for 50 epochs, with batch size 64 (for both the training and validation sets). The initial number of channels was set to 16. Each sub-supernet took 5-20 epochs to converge. We used the momentum SGD optimizer with an initial learning rate of 0.025, followed by a cosine learning rate schedule through the training. We used a momentum rate of 0.9 and a weight decay of $3 \times 10^{-4}$. This experiment ran on 10 P100 GPUs for training both supernet and sub-supernets.

We trained the network for 1500 epochs using a batch size of 128 and use a momentum SGD optimizer with an initial learning rate of 0.025, followed by a cosine learning rate schedule through the training. We use weight decay as the regularization. 

\paragraph{Search Setup for DARTs on PTB.} The search space and the training setup of (sub)supernets are identical to DARTS~\cite{DARTS}. Concretely, both the embedding and the hidden sizes were set to 300. We used 6 P100 GPUs to train both the supernet and 5 (sub)supernets. Each (sub)supernet was trained for 50 epochs using SGD without momentum, with a learning rate of 20. The batch size was set to 256, and the weight decay was set to $3 \times 5^{-7}$. We applied a variational dropout of 0.2 to word embeddings, 0.75 to the cell input, and 0.25 to all the hidden nodes. We also applied a dropout rate of 0.75 to the output layer.

\paragraph{Search Setup for ImageNet~\cite{autogan}.} For proxylessNas, we exactly keep the same search pipeline with original paper~\cite{proxylessnas}. We randomly sample 50,000 images from the training set as a validation set during the architecture search. For our few-shot NAS, we split 3 sub-supernets. The (sub)supernets parameters are updated using the Adam optimizer with an initial learning rate of 0.001. The (sub)supernets are trained on the remaining training images with batch size 256. For once for all NAS, the search setup is also consistent with the original OFA~\cite{ofa}. In specific, we use the same architecture space as MobileNetV3~\cite{mobilenetv3}; for supernet training, we use the standard SGD optimizer with Nesterov momentum 0.9, and weight decay is set to $3 \times 10^{-5}$. The initial learning rate is 2.6, and we use the cosine schedule for learning rate decay. We split 5 sub-supernets. The (sub)supernets are trained for 180 epochs with batch size 2048 on 64 32G V100 GPUs.

\paragraph{Search Setup for AutoGAN~\cite{autogan}.} Our search and training settings were identical to AutoGAN~\cite{autogan}, which followed spectral normalization GAN~\cite{gansetup} when training the (sub-)-supernets. We split the supernet (shared GAN in~\cite{autogan}) into 3 sub-supernets. The learning rate of both generator and discriminator was set to $2e^{-4}$. We used the hinge loss and an Adam optimizer. The batch size of the discriminator was 64, and the generator was 128. The initial learning rate was set to $3.5e^{-4}$. The AutoGAN searched for 90 iterations for one supernet. For each iteration, the shared GAN (supernet) was trained for 15 epochs, and the controller was trained for 30 steps. After the shared GAN (supernet) was trained, we transferred the weight to each sub-supernets and trained them for 12 epochs. We trained the controller with 30 steps. The discovered architectures were trained for 50,000 generator iterations. We used 4 P100 GPUs in this experiment.

\section{One-shot NAS v.s. Few-shot NAS by Robust DARTS ~\cite{robustdarts}}

        \makeatletter\def\@captype{table}\makeatother\caption{Few-shot Robust DARTS vs. One-shot Robust DARTS over 4 Search Space}
        \label{robustdarts_table}
\begin{center}

        \footnotesize{
\begin{tabular}{ccc}
\hline
\textbf{Method}                         & \textbf{Space}          & \textbf{Top 1 Acc(\%)}              \\ \hline
\multicolumn{1}{|c|}{one-shot}          & \multicolumn{1}{c|}{s1} & \multicolumn{1}{c|}{96.49}          \\
\multicolumn{1}{|c|}{\textbf{few-shot}} & \multicolumn{1}{c|}{s1} & \multicolumn{1}{c|}{\textbf{96.81}} \\ \hline
\multicolumn{1}{|c|}{one-shot}          & \multicolumn{1}{c|}{s2} & \multicolumn{1}{c|}{96.22}          \\
\multicolumn{1}{|c|}{\textbf{few-shot}} & \multicolumn{1}{c|}{s2} & \multicolumn{1}{c|}{\textbf{96.55}} \\ \hline
\multicolumn{1}{|c|}{one-shot}          & \multicolumn{1}{c|}{s3} & \multicolumn{1}{c|}{97.19}          \\
\multicolumn{1}{|c|}{\textbf{few-shot}} & \multicolumn{1}{c|}{s3} & \multicolumn{1}{c|}{\textbf{97.28}} \\ \hline
\multicolumn{1}{|c|}{one-shot}          & \multicolumn{1}{c|}{s4} & \multicolumn{1}{c|}{95.60}          \\
\multicolumn{1}{|c|}{\textbf{few-shot}} & \multicolumn{1}{c|}{s4} & \multicolumn{1}{c|}{\textbf{96.30}} \\ \hline
\end{tabular}%

}
\end{center}

We use our few-shot NAS with Robust DARTS searching architectures over 4 different search spaces defined by the original paper ~\cite{robustdarts}. For table ~\ref{robustdarts_table}, we can see that the accuracy of architectures searched by our few-shot is significantly better than one-shot over all 4 search spaces. Our training setup is strictly the same as its original paper.

\end{document}